\documentclass{article}
\usepackage{iclr2027_conference,times}
\usepackage[T1]{fontenc}
\usepackage{amsmath,amssymb,booktabs,graphicx}
\usepackage{xcolor,tikz,hyperref,url,flafter}
\usetikzlibrary{arrows.meta,positioning,fit}
\hypersetup{colorlinks=true,linkcolor=blue!45!black,citecolor=blue!45!black,urlcolor=blue!45!black,
pdftitle={From Target Selection to Condition Reuse},pdfauthor={Tianda Sun and Dimitar Kazakov}}
\definecolor{randomblue}{HTML}{246B9B}
\definecolor{pathorange}{HTML}{BC632A}
\definecolor{textteal}{HTML}{238572}
\definecolor{naturalpurple}{HTML}{8063A9}
\definecolor{nativegray}{HTML}{777777}
\newcommand{\arm}[1]{\textsc{#1}}

\newcommand{\RandomDose}{0.877}

\newcommand{\PathDose}{0.870}
\newcommand{\ReaderBA}{0.933}
\newcommand{\ReaderPermutationBA}{0.493}

\newcommand{\FreshRandomAnchorNativeSelect}{35/128}

\newcommand{\FreshRandomAnchorWriterSelect}{119/128}

\newcommand{\FreshPathAnchorNativeSelect}{33/128}

\newcommand{\FreshPathAnchorWriterSelect}{114/128}

\title{From Target Selection to Condition Reuse:\\Auditing Residual Interventions in Tool-Use Models}
\author{Tianda Sun \quad Dimitar Kazakov\\University of York\\\textnormal{\small\url{https://github.com/TiandaSun/residual-condition-reuse}}}
\iclrfinalcopy
\begin{document}
\maketitle
\lhead{Preprint}
\begin{abstract}
Activation interventions can make a language model produce a target answer, but can the model use the intended condition in other computations? We study this question in Qwen3.5-27B on tool-use tasks where time or permissions determine eligibility and selection. Across 32 fresh worlds, two case-specific residual-writing variants raise familiar target-selection correctness from 35/128 to 119/128 and from 33/128 to 114/128. Yet these gains disappear when familiar facts and identifiers are recombined within the same selection task, while selection of previously rewarded identifiers increases. The gains persist across output formats but yield little improvement in independent judgments about whether a candidate is eligible or should be selected. Internal edits can nevertheless change how a source fact affects a judgment: in a prespecified secondary comparison, an early-layer source patch changes 50/64 judgments when that source is used, with one adverse change among 64 cases where it is not. This selective effect does not ensure correct complete actions. Together, these results separate target-selection control, condition-sensitive judgments, and complete-action correctness. Evaluating how an edited state is used, rather than only its trained answer, exposes these differences.

\end{abstract}
\section{Introduction}
Activation interventions can transfer task behavior to new contexts \citep{todd2024function}. But a desired answer leaves a further question: \emph{can the model use the intended change in other computations?} An edit may make a model select a service under a different permission profile. Does the same profile guide its judgments when facts are rearranged or eligibility is queried directly?

Recent work distinguishes task control from format-invariant representations \citep{opielka2026causality}, attributes steering effects to answer identifiers versus semantic labels \citep{gao2026control}, and tests reusable causal interfaces \citep{ma2026hidden}. We examine what successful target selection establishes when other decisions depend on the same earlier condition.

Our tasks ask for the lowest-cost eligible option under explicit time or permission rules. We train a residual vector for each case on target identifiers. Tests rearrange familiar facts and identifiers within the same selection interface or independently ask whether a candidate is eligible or should be selected. Each query restarts from the edited history without receiving the selection output. Textual updates and natural counterfactual histories provide capability controls.

On 32 fresh worlds, both writer variants strongly improve familiar selection. Factual recombination removes the gain and increases trained-ID repetition, even among fully fitted writers. Selection gains persist across output formats but transfer little to independent Boolean judgments. This selection--verification gap also occurs in post hoc cases with correct matched controls.

Substituting natural counterfactual history changes eligibility judgments, but combining it with learned state retains crossed-selection bias. In a prespecified secondary comparison, an early source patch changes judgments mainly when the edited profile is active, without ensuring a correct full action. These findings distinguish target-selection control, condition-sensitive judgments, and complete-action correctness.

\section{Controlled setting and evaluation targets}
\label{sec:setting}
Consider an episode where a profile determines later eligibility. A service unavailable under profile A becomes the cheapest eligible option under B. Making the model select it controls an answer. Rearranging service facts or independently querying eligibility tests whether the intervention also supports using B. The governing rule gives explicit answers, avoiding subjective judgments about open-ended plans.

\paragraph{Episodes and ground truth.}
Each procedural world specifies two conditions, a canonical initial action valid under both, and future menus. For each evaluation direction, A is the original condition and B is the counterfactual target; the labels reverse in the opposite direction. The initial action is supplied as part of the history. It is not a corrected decision generated by the model. The future task is to produce a \texttt{select\_followup} tool call containing a dependent candidate, an independent offer, and a reservation identifier.

A menu $m$ contains eight dependent candidates and four independent offers. Let $e_c(j,m)$ denote candidate $j$'s eligibility under condition $c$. The dependent target is
\begin{equation}
 y(c,m)=\operatorname*{arg\,min}_{j:e_c(j,m)=1}\operatorname{cost}(j,m).
 \label{eq:winner}
\end{equation}
The construction rules ensure a unique winner. Time rules use a referenced arrival time and a minimum connection interval. Permission rules require every tag requested by a candidate to occur in the referenced profile's grants. A separate capacity-and-cost rule determines the independent offer. We use two construction patterns per rule family; these procedural episodes isolate conditional selection rather than represent the distribution of deployed-agent tasks.

\begin{figure}[htbp]
\centering
\begin{tikzpicture}[font=\normalsize,>=Latex,
card/.style={draw=black!23,line width=.5pt,rounded corners=3pt,
  align=left,inner xsep=6pt,inner ysep=6pt},
consumer/.style={card,anchor=north west,text width=4.15cm,fill=randomblue!3},
link/.style={line width=.7pt,draw=randomblue!75!black},
arrow/.style={link,->}]
\node[card,anchor=west,text width=3.15cm,fill=black!2] (history) at (0,-1.65)
  {\textbf{Initial input: A}\newline Permissions: $p$ only\newline Supplied valid action};
\node[card,anchor=west,text width=2.9cm,draw=randomblue!65,fill=randomblue!10]
  (edit) at (4.02,-1.65)
  {\textbf{One residual write}\newline Block 60 output\newline Save Handoff};
\draw[arrow] (history.east) -- (edit.west);

\node[consumer] (select) at (8.40,0)
  {\textbf{\textcolor{randomblue!80!black}{Select}}\newline Writer chooses $s$\newline Native chooses another ID};
\node[consumer] (changed) at (8.40,-1.73)
  {\textbf{\textcolor{randomblue!80!black}{Changed query about $s$}}\newline Writer: false\newline Text / Natural B: true};
\node[consumer] (stable) at (8.40,-3.46)
  {\textbf{\textcolor{randomblue!80!black}{Stable query}}\newline Same truth in A and B};

\coordinate (fork) at (7.88,-1.65);
\draw[link] (edit.east) -- (fork);
\draw[link] (fork |- select.west) -- (fork |- stable.west);
\fill[randomblue!75!black] (fork) circle (1.5pt);
\foreach \branch in {select,changed,stable}
  \draw[arrow] (fork |- \branch.west) -- (\branch.west);

\node[card,anchor=north west,draw=none,fill=black!3,text width=7cm]
  at (0,-2.70)
  {\textbf{Target B:} $p,q$ (not in the initial input).\newline Future facts arrive in each branch.\newline Example: $s$ costs 33 and requires $p,q$.\newline Under B, $s$ is the cheapest eligible candidate.};
\end{tikzpicture}
\caption{\textbf{A shared intervention, three independent continuations.} The schematic permission example is an excerpt from a verified case, with readable aliases. The writer changes the selected candidate to $s$; a separately restarted eligibility query about $s$ still answers false. Text update and Natural B answer true. No selected output enters either query. The diagram specifies the experimental protocol, not an inferred neural dependency graph.}
\label{fig:protocol}
\end{figure}
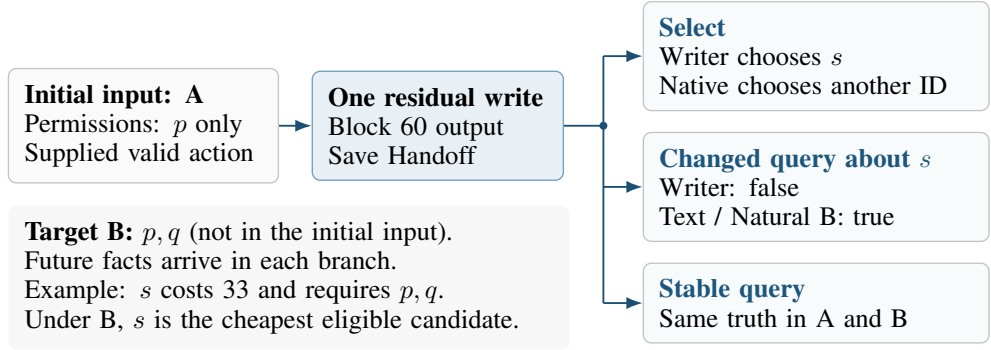

\paragraph{Baseline consumers of the same retained state.}
After prefilling the initial history, we save a \arm{Handoff} state. Each consumer independently resumes a copy of it and receives the future facts. \arm{Select} requests the complete tool action. \arm{Changed} asks for a Boolean JSON eligibility judgment about a candidate whose truth differs between A and B. \arm{Stable} asks about a candidate whose truth does not change. A queried candidate need not be the winner; aggregate disagreement between selection and judgment is therefore not a count of selected candidates judged ineligible. Figure~\ref{fig:protocol} illustrates a verified case where the entities do coincide.

We distinguish \emph{target selection}, which requires valid tool-call syntax and the correct dependent ID, from \emph{full-action correctness}, which also checks the independent offer and reservation. Query correctness is scored against the requested condition, B. Native A can follow its original instructions correctly while scoring poorly against B; these scores measure a counterfactual change objective, not the frequency of natural mistakes.

\paragraph{Familiar menus and factual crossings.}
Each writer is trained on eight menu views. \arm{Anchor} evaluation preserves a training menu. \arm{Cross} combines factual assignments from another training view with the source identifiers and order. The source's trained ID, A's current winner, and B's current winner must all differ. This construction tests whether control follows the recombined facts within the \emph{same selection interface}, rather than simply recovering a previously rewarded identifier. The facts and IDs have been encountered separately; Cross is not a new domain. We use \emph{condition reuse} to mean target-consistent behavior under these recombinations and independent queries, not a pre-assumed internal symbolic variable.

\section{Interventions and experimental design}
\label{sec:methods}
\paragraph{A fixed-site, case-specific writer.}
We freeze Qwen3.5-27B \citep{qwen2026qwen35} and learn a 5,120-dimensional additive residual vector, called a \emph{writer}, for each world, condition direction, and supervision arm. We apply it once at the shared \texttt{CX} prefix token of the candidate identifier in the supplied initial action. The write occurs at the output of zero-based block 60 of 64, after that block has updated its cache. It leaves the already-computed cache in blocks 0--60 unchanged, while the altered residual can change the retained state produced by blocks 61--63. These retained states form the Handoff used by the independent consumers.

The model prefills the entire supplied initial history. Future facts arrive only after the write, and the hook is removed for all future computation. Inference uses BF16, greedy no-thinking generation, and a 1,024-token output cap. The loss averages teacher-forced cross-entropy over eight training views ($1/8$) and the 13 tokens of each target identifier ($1/13$). Writing $a_v=y(B,m_v)$ for the target in view $v$, we optimize
\begin{equation}
\mathcal{L}(\delta)=\frac{1}{8}\sum_{v=1}^{8}\frac{1}{13}\sum_{t=1}^{13}
-\log p_\theta(a_{v,t}\mid h_A,m_v,a_{v,<t};\delta),\qquad \|\delta\|_2\leq R.
\label{eq:objective}
\end{equation}
Here $h_A$ is the initial history; the context also includes the fixed tool-call wrapper preceding the ID. One vector serves all views of that world and direction. Judgments, independent fields, and probe outputs receive no supervision. We use 96 projected normalized-gradient updates from zero, with fixed $R=401.26$, and evaluate the final vector (Appendix~\ref{app:runtime}). Behavioral results use free generation.

\paragraph{Two factual-supervision variants.}
\arm{Random-fact} uses four joint permutations of cost and eligibility facts, each with two ID/order views. \arm{Path-cover} uses two cost-only and two eligibility-only permutations, again with two ID/order views. This is dependency-guided example selection, not injection of a graph embedding. Each variant uses its own set of menus, called its \emph{bank}; its controls use the same menus. Earlier label-only and repeated-view comparisons appear in Appendix~\ref{app:discovery}.

\paragraph{Capability and collateral-behavior controls.}
\arm{Native A} retains the original history. \arm{Text update} explicitly states the target condition at the consumer, before future facts. \arm{Natural B} builds the target history from the start. Text and Natural B test whether the model can express the target behavior with more direct support; their access to the condition is not matched to a single-site residual write. Full-action and Stable-query outcomes track behavior outside the supervised identifier.

\paragraph{Confirmation and exploratory cohorts.}
The primary results use 32 fresh worlds, eight per construction pattern, generated after the protocol was frozen. Writers are fitted anew: $32\times2$ directions $\times2$ arms gives 128 vectors. We evaluate two source menus per direction and arm, giving 128 menus per arm in each evaluation family (Anchor or Cross), or 512 across both arms and families. All directions and fits are retained, regardless of optimization success or evaluation outcome. This confirms the \emph{fitting procedure} on new worlds; it does not transfer one vector to 32 new tasks. Sixteen earlier worlds informed the protocol and support a separate exploratory composition test.

\paragraph{Primary contrasts and uncertainty.}
For each world $u$ and writer arm $w$, let $S_{w,u}$ be the mean writer-minus-Native selection difference over its four Anchor menus, and $Q_{w,u}$ the matched Changed-query difference. We report both selection gain $S$ and its separation from query gain, $D=S-Q$, for each arm. A fifth contrast, $K$, tests the effect of target history on the original queries (Section~\ref{sec:history}).

We jointly bootstrap worlds within the four fixed pattern strata, using 100,000 draws. The five registered contrasts receive 99\% percentile intervals, giving approximately 95\% simultaneous coverage under a Bonferroni correction. Worlds, not generations or candidate rows, are the independent units. Cross counts, fit-conditioned subsets, historical diagnostics, and the old-world composition are descriptive or exploratory.

\paragraph{Frozen-writer interface tests.}
We reuse all 32 confirmation worlds and 128 fitted writers on Anchor and Cross without retraining. In addition to the original selection interface, we test two fixed wordings in each of two formats: plain JSON and a tool wrapper. This gives five selection variants. Winner and eligibility questions use the same four wording--format combinations, giving eight Boolean questions per candidate. Within each wording, changing the wrapper leaves the question fixed.

The primary Boolean comparison uses Anchor candidates whose winner and eligibility labels agree \emph{within} each condition, with both labels flipping from A to B. All 128 Anchor menus per bank contribute. Correctness requires the right candidate ID and truth value. We average across candidates and wordings within menus, then across menus within worlds. Eight registered effects compare winner gain and winner-minus-eligibility gain across two arms and two Boolean formats, using joint stratified 99.375\% bootstrap intervals. Selection, paired selection-minus-winner, and wrapper contrasts receive descriptive 95\% intervals (Appendix~\ref{app:interfaces}).

\section{Selection control does not establish condition reuse}
\label{sec:results}
\subsection{Familiar target selections are strongly controllable}
\begin{figure}[tb]
\centering
\begin{tikzpicture}[x=1cm,y=1.15cm,font=\normalsize]
\begin{scope}[shift={(0.0,0)}]
\node[anchor=west,font=\normalsize\bfseries] at (0,4.7) {(a) Familiar selection};
\draw[gray!22] (1.25,.45) -- (1.25,4.35);
\node[below,font=\normalsize] at (1.25,.45) {0};
\draw[gray!22] (3.0,.45) -- (3.0,4.35);
\node[below,font=\normalsize] at (3.0,.45) {50};
\draw[gray!22] (4.75,.45) -- (4.75,4.35);
\node[below,font=\normalsize] at (4.75,.45) {100};
\node[anchor=west,font=\normalsize\bfseries] at (0,4.2) {Random-fact};
\node[anchor=east,font=\normalsize] at (1.16,3.7699999999999996) {Native};
\draw[nativegray,line width=4pt] (1.25,3.7699999999999996) -- (2.2070,3.7699999999999996);
\node[anchor=west,font=\normalsize] at (2.2670,3.7699999999999996) {35/128};
\node[anchor=east,font=\normalsize] at (1.16,3.4299999999999997) {Writer};
\draw[randomblue,line width=4pt] (1.25,3.4299999999999997) -- (4.5039,3.4299999999999997);
\node[anchor=west,font=\normalsize] at (4.5639,3.4299999999999997) {119/128};
\node[anchor=east,font=\normalsize] at (1.16,3.0899999999999994) {Text};
\draw[textteal,line width=4pt] (1.25,3.0899999999999994) -- (3.0273,3.0899999999999994);
\node[anchor=west,font=\normalsize] at (3.0873,3.0899999999999994) {65/128};
\node[anchor=east,font=\normalsize] at (1.16,2.7499999999999996) {Natural B};
\draw[naturalpurple,line width=4pt] (1.25,2.7499999999999996) -- (2.7539,2.7499999999999996);
\node[anchor=west,font=\normalsize] at (2.8139,2.7499999999999996) {55/128};
\node[anchor=west,font=\normalsize\bfseries] at (0,2.3) {Path-cover};
\node[anchor=east,font=\normalsize] at (1.16,1.8699999999999999) {Native};
\draw[nativegray,line width=4pt] (1.25,1.8699999999999999) -- (2.1523,1.8699999999999999);
\node[anchor=west,font=\normalsize] at (2.2123,1.8699999999999999) {33/128};
\node[anchor=east,font=\normalsize] at (1.16,1.5299999999999998) {Writer};
\draw[pathorange,line width=4pt] (1.25,1.5299999999999998) -- (4.3672,1.5299999999999998);
\node[anchor=west,font=\normalsize] at (4.4272,1.5299999999999998) {114/128};
\node[anchor=east,font=\normalsize] at (1.16,1.19) {Text};
\draw[textteal,line width=4pt] (1.25,1.19) -- (3.0000,1.19);
\node[anchor=west,font=\normalsize] at (3.0600,1.19) {64/128};
\node[anchor=east,font=\normalsize] at (1.16,0.8499999999999999) {Natural B};
\draw[naturalpurple,line width=4pt] (1.25,0.8499999999999999) -- (2.8359,0.8499999999999999);
\node[anchor=west,font=\normalsize] at (2.8959,0.8499999999999999) {58/128};
\node[font=\normalsize] at (3,-.2) {Target-selection correctness (\%)};
\end{scope}
\begin{scope}[shift={(6.25,0)}]
\node[anchor=west,font=\normalsize\bfseries] at (0,4.7) {(b) Crossed-fact selection};
\draw[gray!22] (1.25,.45) -- (1.25,4.35);
\node[below,font=\normalsize] at (1.25,.45) {0};
\draw[gray!22] (3.0,.45) -- (3.0,4.35);
\node[below,font=\normalsize] at (3.0,.45) {50};
\draw[gray!22] (4.75,.45) -- (4.75,4.35);
\node[below,font=\normalsize] at (4.75,.45) {100};
\node[anchor=west,font=\normalsize\bfseries] at (0,4.2) {Random-fact};
\node[anchor=east,font=\normalsize] at (1.16,3.7699999999999996) {Native};
\draw[nativegray,line width=4pt] (1.25,3.7699999999999996) -- (2.2617,3.7699999999999996);
\node[anchor=west,font=\normalsize] at (2.3217,3.7699999999999996) {37/128};
\node[anchor=east,font=\normalsize] at (1.16,3.4299999999999997) {Writer};
\draw[randomblue,line width=4pt] (1.25,3.4299999999999997) -- (2.0703,3.4299999999999997);
\node[anchor=west,font=\normalsize] at (2.1303,3.4299999999999997) {30/128};
\node[anchor=east,font=\normalsize] at (1.16,3.0899999999999994) {Text};
\draw[textteal,line width=4pt] (1.25,3.0899999999999994) -- (3.0273,3.0899999999999994);
\node[anchor=west,font=\normalsize] at (3.0873,3.0899999999999994) {65/128};
\node[anchor=east,font=\normalsize] at (1.16,2.7499999999999996) {Natural B};
\draw[naturalpurple,line width=4pt] (1.25,2.7499999999999996) -- (2.9727,2.7499999999999996);
\node[anchor=west,font=\normalsize] at (3.0327,2.7499999999999996) {63/128};
\node[anchor=west,font=\normalsize\bfseries] at (0,2.3) {Path-cover};
\node[anchor=east,font=\normalsize] at (1.16,1.8699999999999999) {Native};
\draw[nativegray,line width=4pt] (1.25,1.8699999999999999) -- (2.2070,1.8699999999999999);
\node[anchor=west,font=\normalsize] at (2.2670,1.8699999999999999) {35/128};
\node[anchor=east,font=\normalsize] at (1.16,1.5299999999999998) {Writer};
\draw[pathorange,line width=4pt] (1.25,1.5299999999999998) -- (2.1523,1.5299999999999998);
\node[anchor=west,font=\normalsize] at (2.2123,1.5299999999999998) {33/128};
\node[anchor=east,font=\normalsize] at (1.16,1.19) {Text};
\draw[textteal,line width=4pt] (1.25,1.19) -- (3.0000,1.19);
\node[anchor=west,font=\normalsize] at (3.0600,1.19) {64/128};
\node[anchor=east,font=\normalsize] at (1.16,0.8499999999999999) {Natural B};
\draw[naturalpurple,line width=4pt] (1.25,0.8499999999999999) -- (2.8359,0.8499999999999999);
\node[anchor=west,font=\normalsize] at (2.8959,0.8499999999999999) {58/128};
\node[font=\normalsize] at (3,-.2) {Target-selection correctness (\%)};
\end{scope}
\node[anchor=west,font=\normalsize\bfseries] at (0,-.85) {(c) Cross: attraction to the old trained ID};
\node[anchor=west] at (0,-1.5) {Random-fact};
\draw[randomblue,line width=1.2pt] (3.8438,-1.5) -- (7.6406,-1.5);
\draw[randomblue,fill=white,thick] (3.8438,-1.5) circle (2.5pt);
\fill[randomblue] (7.6406,-1.5) circle (2.5pt);
\node[anchor=east,font=\normalsize] at (3.7237,-1.5) {6/128};
\node[anchor=west,font=\normalsize] at (7.7606,-1.5) {33/128};
\node[anchor=west] at (0,-2.05) {Path-cover};
\draw[pathorange,line width=1.2pt] (3.5625,-2.05) -- (5.8125,-2.05);
\draw[pathorange,fill=white,thick] (3.5625,-2.05) circle (2.5pt);
\fill[pathorange] (5.8125,-2.05) circle (2.5pt);
\node[anchor=east,font=\normalsize] at (3.4425,-2.05) {4/128};
\node[anchor=west,font=\normalsize] at (5.9325,-2.05) {20/128};
\node[anchor=west,font=\normalsize,align=left] at (9.1,-1.8) {Open: Native A\\Filled: Writer};
\draw[->,gray] (3,-2.5) -- (8.65,-2.5);
\node[below,font=\normalsize] at (3.0,-2.5) {0};
\node[below,font=\normalsize] at (4.8,-2.5) {10};
\node[below,font=\normalsize] at (6.6,-2.5) {20};
\node[below,font=\normalsize] at (8.399999999999999,-2.5) {30};
\node[font=\normalsize] at (5.7,-3.15) {Old-ID selection (\%)};
\end{tikzpicture}
\caption{\textbf{The selection gain disappears when familiar facts and identifiers are recombined.} Counts are out of 128 menus per arm and family, nested in 32 fresh worlds. Anchor menus are training-exposed; Cross preserves the selection interface but changes the fact--ID association. The lower panel shows increased selection of the old trained ID, which is never the current Cross target. Controls are paired within bank. Raw bars are descriptive; clustered intervals for the registered Anchor gains are reported in the text.}
\label{fig:selection}
\end{figure}
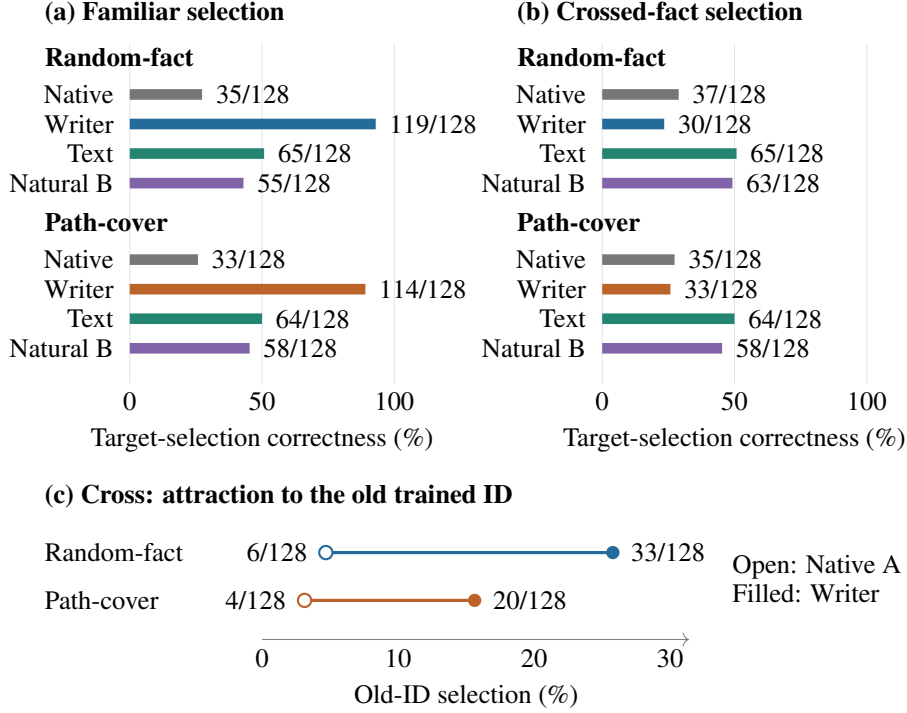

On Anchor menus, Random-fact increases target selection from \FreshRandomAnchorNativeSelect{} to \FreshRandomAnchorWriterSelect{}; Path-cover increases it from \FreshPathAnchorNativeSelect{} to \FreshPathAnchorWriterSelect{} (Figure~\ref{fig:selection}). Both arms have positive mean selection gains in every one of the 32 fresh worlds. The world-clustered gains are 65.63 percentage points (99\% interval [57.03, 74.22]) and 63.28 points ([53.13, 72.66]), respectively.

The complete tool action is less controllable than its target field (Table~\ref{tab:anchor}). Full-action correctness reaches 68/128 and 64/128, compared with 119/128 and 114/128 target selections. Some of this gap precedes intervention: Native A already has incorrect independent fields. Relative to Native A, the independent offer has seven gains and twelve losses under Random-fact, and six gains and seven losses under Path-cover. Target-ID success is therefore neither complete-action success nor a guarantee of unchanged collateral behavior.

\begin{table}[htbp]
\caption{\textbf{Familiar selection, complete actions, and independent queries.} All counts are out of 128 own-bank Anchor menus from 32 fresh worlds. Select checks schema and the target-dependent ID; Full additionally checks the independent offer and reservation. Changed and Stable are separate eligibility branches, scored under B. Writer query responses are token-identical to Native A in every matched case. Text and Natural B receive the target condition through different, broader support.}
\label{tab:anchor}
\centering\small
\begin{tabular}{llrrrr}
\toprule
Bank & Arm & Select & Full & Changed & Stable \\
\midrule
Random-fact & Native A & 35/128 & 24/128 & 24/128 & 123/128 \\
Random-fact & Writer & 119/128 & 68/128 & 24/128 & 123/128 \\
Random-fact & Text update & 65/128 & 39/128 & 96/128 & 123/128 \\
Random-fact & Natural B & 55/128 & 32/128 & 103/128 & 123/128 \\
Path-cover & Native A & 33/128 & 18/128 & 23/128 & 126/128 \\
Path-cover & Writer & 114/128 & 64/128 & 23/128 & 126/128 \\
Path-cover & Text update & 64/128 & 38/128 & 99/128 & 124/128 \\
Path-cover & Natural B & 58/128 & 32/128 & 102/128 & 123/128 \\
\bottomrule
\end{tabular}

\end{table}

\subsection{Crossed facts expose attraction to trained identifiers}
Cross changes what the familiar identifiers denote while retaining the same tool-call task. Target-selection correctness falls from 37/128 with Native A to 30/128 with the Random-fact writer, and from 35/128 to 33/128 with Path-cover.

Direct access to the target condition improves selection on these same crossed menus: Text update reaches 65/128 and 64/128, and Natural B reaches 63/128 and 58/128. Selection remains imperfect under these controls, but correct recomputation is substantially more frequent than under the writer.

The edits also increase selection of the \emph{old trained ID}: from 6/128 to 33/128 for Random-fact, and from 4/128 to 20/128 for Path-cover. By construction, that identifier is not the current counterfactual winner. Sixteen of the 33 Random-fact repetitions and six of the 20 Path-cover repetitions are also ineligible under B. Thus factual crossing reveals a specific directional error, not merely a loss of accuracy: the intervention more often recovers a formerly rewarded identifier despite changed facts.

\paragraph{A concrete crossing.}
In one recorded permission example, the source menu's target under B is candidate $r$, costing 24. Cross retains the IDs and order but reassigns the observed facts. Candidate $r$ now costs 41 and requires two permissions absent from B. Candidate $s$, costing 78, is eligible and becomes B's unique cheapest choice; A's current winner is a third ID. The frozen writer nevertheless emits $r$ in a valid call, with the independent offer and reservation correct. This post hoc excerpt illustrates how retaining a rewarded ID differs from applying the condition to current facts; full identifiers are in Appendix~\ref{app:example}.

\paragraph{Does this only reflect incomplete training fit?}
Forty-four of 64 Random-fact writers and 39 of 64 Path-cover writers fit all eight dependent-ID training targets. In this post hoc subset, Cross target correctness still changes from 37 to 25 out of 88 menus, and from 32 to 29 out of 78. Old-ID selections rise from 4 to 24 and from 3 to 13. The main analysis retains every writer; this fit-conditioned check shows that the observed failure is not confined to vectors that never learned their supervised targets.

\subsection{Original eligibility responses remain unchanged}
Table~\ref{tab:anchor} reports eligibility queries independently restarted from the same retained state. On Anchor, Changed correctness is 24/128 for both Native A and Random-fact, and 23/128 for both Native A and Path-cover. More strongly, every matched Changed response is token-identical to Native A. All matched Stable responses are token-identical as well.

In contrast, Text update answers Changed correctly on 96/128 and 99/128 cases; Natural B answers 103/128 and 102/128. These controls establish that the queried condition can affect this interface. Because the observed query gain is zero, the two registered separation contrasts $D=S-Q$ coincide with $S$ in this sample, including the 99\% intervals in Section~\ref{sec:results}.

We next vary the selection wrapper and ask matched winner-verification questions to examine how the requested task affects this separation.

\subsection{Selection gains persist across wrappers but not verification}
\label{sec:interfaces}
Could the selection/query separation simply reflect tool formatting? With the writers fixed, plain-JSON and tool-wrapped selection retain 57.42--62.50 percentage-point gains on Anchor (Table~\ref{tab:interfaces}). Each fixed wording supports the effect. Tool-minus-JSON gain is 1.95 points for Random-fact (95\% interval [-2.34, 6.25]) and 0 for Path-cover ([-3.52, 3.13]). These wrapper changes do not account for the much larger attenuation when the requested output becomes a Boolean judgment.

Winner verification asks whether a candidate is the lowest-cost eligible option, rather than merely eligible. On candidates matched across these two questions, winner gain ranges from -1.17 to 0 points and eligibility gain from 0 to 0.39. The gain on original selection exceeds the winner-verification gain by 63.67--66.02 points; all descriptive 95\% lower bounds for these paired differences are at least 56.25. None of the four writer/interface combinations has positive adjusted lower bounds for both winner gain and winner-minus-eligibility gain. Thus asking whether an option is the winner also fails to inherit the large benefit seen when selecting it.

\begin{table}[htbp]
\caption{\textbf{Large selection gains, weak verification gains, and asymmetric natural capability.} Top: writer-minus-Native Anchor effects (percentage points), with equal weight on the same 32 worlds. New interfaces average two fixed wordings. Bottom: Natural B correctness on matched JSON questions, split by target truth; these are query-weighted counts, not the primary estimand.}
\label{tab:interfaces}
\centering\small
\begin{tabular}{llrr}
\toprule
Task & Format & Random-fact & Path-cover \\
\midrule
Selection & Original & 65.63 & 63.28 \\
Selection & JSON & 60.55 & 57.42 \\
Selection & Tool & 62.50 & 57.42 \\
Winner & JSON & -0.39 & -1.17 \\
Winner & Tool & 0.00 & -0.39 \\
Eligibility & JSON & 0.00 & 0.00 \\
Eligibility & Tool & 0.39 & 0.00 \\
\midrule Bank & Judgment & B false & B true \\
Random-fact & Winner & 142/154 & 52/158 \\
Random-fact & Eligibility & 113/154 & 148/158 \\
Path-cover & Winner & 145/160 & 55/160 \\
Path-cover & Eligibility & 117/160 & 148/160 \\
\bottomrule
\end{tabular}

\end{table}

\paragraph{Capability qualifies the verification result.}
Natural B answers only 33--34\% of target-true winner questions correctly, versus 91--92\% of target-false questions; pooled accuracy hides this asymmetry (Table~\ref{tab:interfaces}). Nevertheless, relative to Native A, Natural B raises matched JSON winner correctness from 106/312 to 194/312 for Random-fact and from 115/320 to 200/320 for Path-cover. These denominators count candidate questions across both wordings, not additional worlds.

The gap also appears when the controls succeed. A post hoc screen finds 27 permission records in 12 worlds, each pairing one fitted writer with one Anchor menu. In every record, all five writer selection variants choose the B target. Native A answers all eight matched Boolean questions correctly under A, and Natural B and Text each answer all eight correctly under B. Yet none of the writer's eight judgments is correct under B. One instance selects a cost-28 service requiring two B-only permissions, while its independent winner and eligibility forks still answer false. Because these records were selected by their outcomes, they do not estimate how often the gap occurs (Appendix~\ref{app:interfaces}).

Factual crossing remains the complementary same-task test: new-wrapper selection gains range from -4.69 to 1.95 points. In JSON selection, old-target responses increase from 13 to 51/256 and from 6 to 33/256 across the two banks. Thus tolerating serialization changes does not imply recomputation under changed facts.

\section{Intervention support and condition use}
\label{sec:history}
The late writer leaves most retained history under condition A. We test broader history substitution, then a source-fact patch whose relevance depends on which profile the task uses.

\subsection{Natural history changes judgments, with rule-dependent effects}
We partition retained cache into lower blocks 0--60 and upper blocks 61--63. In AA, AB, BA and BB, the first letter identifies the lower donor condition and the second the upper. We replace cached tensors and continuation metadata, discard stale Handoff logits, and resume on the same nonempty future suffix without replaying earlier history. All blocks execute on future tokens. The lower group contains 95.32\% of retained bytes, so this is a broad history intervention.

\begin{figure}[htbp]
\centering
\begin{tikzpicture}[x=1cm,y=1cm,font=\normalsize]
\node[anchor=west,font=\normalsize] at (0,3.65) {Lower 0--60 (95.32\% bytes)};
\node[anchor=west,font=\normalsize] at (0,3.2) {Upper 61--63};
\node[font=\bfseries] at (5.0,4.18) {AA};
\node[draw=nativegray,fill=nativegray!12,minimum width=1.3cm,minimum height=.37cm,font=\normalsize] at (5.0,3.65) {Donor A};
\node[draw=nativegray,fill=nativegray!12,minimum width=1.3cm,minimum height=.37cm,font=\normalsize] at (5.0,3.2) {Donor A};
\node[font=\bfseries] at (7.0,4.18) {AB};
\node[draw=nativegray,fill=nativegray!12,minimum width=1.3cm,minimum height=.37cm,font=\normalsize] at (7.0,3.65) {Donor A};
\node[draw=textteal,fill=textteal!12,minimum width=1.3cm,minimum height=.37cm,font=\normalsize] at (7.0,3.2) {Donor B};
\node[font=\bfseries] at (9.0,4.18) {BA};
\node[draw=textteal,fill=textteal!12,minimum width=1.3cm,minimum height=.37cm,font=\normalsize] at (9.0,3.65) {Donor B};
\node[draw=nativegray,fill=nativegray!12,minimum width=1.3cm,minimum height=.37cm,font=\normalsize] at (9.0,3.2) {Donor A};
\node[font=\bfseries] at (11.0,4.18) {BB};
\node[draw=textteal,fill=textteal!12,minimum width=1.3cm,minimum height=.37cm,font=\normalsize] at (11.0,3.65) {Donor B};
\node[draw=textteal,fill=textteal!12,minimum width=1.3cm,minimum height=.37cm,font=\normalsize] at (11.0,3.2) {Donor B};
\begin{scope}[shift={(0.0,0)}]
\node[anchor=west,font=\bfseries] at (0,2.55) {Permission};
\draw[gray!22] (0.8,.3) -- (0.8,2.2);
\node[below,font=\normalsize] at (0.8,.3) {0};
\draw[gray!22] (2.5999999999999996,.3) -- (2.5999999999999996,2.2);
\node[below,font=\normalsize] at (2.5999999999999996,.3) {50};
\draw[gray!22] (4.3999999999999995,.3) -- (4.3999999999999995,2.2);
\node[below,font=\normalsize] at (4.3999999999999995,.3) {100};
\node[anchor=east] at (.65,2.02) {AA};
\draw[nativegray,line width=5pt] (.8,2.02) -- (0.8000,2.02);
\node[anchor=west,font=\normalsize] at (0.8600,2.02) {0/256};
\node[anchor=east] at (.65,1.59) {AB};
\draw[naturalpurple,line width=5pt] (.8,1.59) -- (0.8000,1.59);
\node[anchor=west,font=\normalsize] at (0.8600,1.59) {0/256};
\node[anchor=east] at (.65,1.1600000000000001) {BA};
\draw[textteal,line width=5pt] (.8,1.1600000000000001) -- (4.4000,1.1600000000000001);
\node[anchor=west,font=\normalsize] at (4.4600,1.1600000000000001) {256/256};
\node[anchor=east] at (.65,0.73) {BB};
\draw[randomblue,line width=5pt] (.8,0.73) -- (4.4000,0.73);
\node[anchor=west,font=\normalsize] at (4.4600,0.73) {256/256};
\node[font=\normalsize] at (2.6,-.45) {Changed-query target correctness (\%)};
\end{scope}
\begin{scope}[shift={(6.25,0)}]
\node[anchor=west,font=\bfseries] at (0,2.55) {Time};
\draw[gray!22] (0.8,.3) -- (0.8,2.2);
\node[below,font=\normalsize] at (0.8,.3) {0};
\draw[gray!22] (2.5999999999999996,.3) -- (2.5999999999999996,2.2);
\node[below,font=\normalsize] at (2.5999999999999996,.3) {50};
\draw[gray!22] (4.3999999999999995,.3) -- (4.3999999999999995,2.2);
\node[below,font=\normalsize] at (4.3999999999999995,.3) {100};
\node[anchor=east] at (.65,2.02) {AA};
\draw[nativegray,line width=5pt] (.8,2.02) -- (2.1219,2.02);
\node[anchor=west,font=\normalsize] at (2.1819,2.02) {94/256};
\node[anchor=east] at (.65,1.59) {AB};
\draw[naturalpurple,line width=5pt] (.8,1.59) -- (2.1219,1.59);
\node[anchor=west,font=\normalsize] at (2.1819,1.59) {94/256};
\node[anchor=east] at (.65,1.1600000000000001) {BA};
\draw[textteal,line width=5pt] (.8,1.1600000000000001) -- (2.9375,1.1600000000000001);
\node[anchor=west,font=\normalsize] at (2.9975,1.1600000000000001) {152/256};
\node[anchor=east] at (.65,0.73) {BB};
\draw[randomblue,line width=5pt] (.8,0.73) -- (2.9234,0.73);
\node[anchor=west,font=\normalsize] at (2.9834,0.73) {151/256};
\node[font=\normalsize] at (2.6,-.45) {Changed-query target correctness (\%)};
\end{scope}
\end{tikzpicture}
\caption{\textbf{Natural history makes the original query condition-responsive.} Each panel contains 256 menus from 16 of the 32 confirmation worlds, pooling both banks and Anchor/Cross menus within a rule family. Cache letters specify lower (0--60; 95.32\% of bytes) and upper (61--63) donors. Permission judgments switch completely; time judgments improve partly.}
\label{fig:cache}
\end{figure}
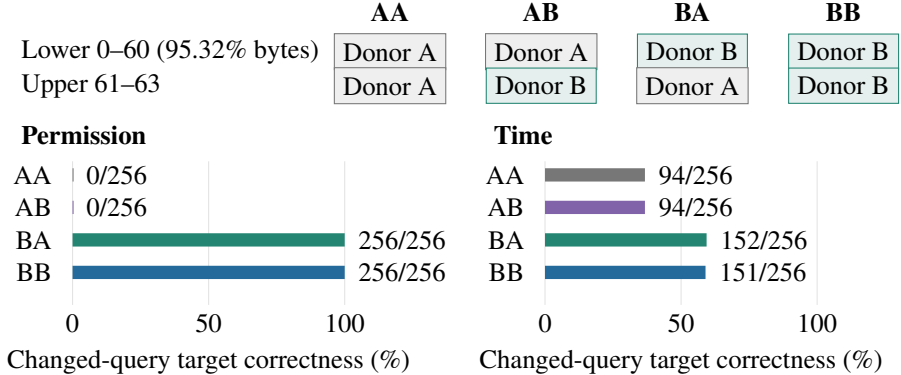

Across 512 menus, Changed correctness is 94, 94, 408 and 407 under AA, AB, BA and BB. The registered history contrast $K$ averages BA-minus-AB over 16 menus per world, then over worlds: +61.33 points (99\% interval [54.88, 68.75]), positive in 25/32 worlds. All seven zero-gain worlds are time cases.

For permission queries, AA follows A in all 256 cases and BA follows B in all 256. Scored against B, this changes correctness from 0/256 to 256/256. Time-query correctness rises from 94/256 to 152/256 and remains imperfect (Figure~\ref{fig:cache}).

AA-to-BA target selection rises from 140 to 235/512 and full-action correctness from 84 to 130/512. Stable correctness falls from 499 to 497/512, with eleven gains and thirteen losses (Appendix~\ref{app:stats}).

\subsection{Source interventions reveal use-dependent factual effects}
\label{sec:source}
Can a source patch change judgments according to whether the task uses the edited fact? We reuse the 16 confirmation permission worlds, each with two profiles. The patch replaces residual rows for one profile's grants with natural fact-counterfactual values, over 51 or 61 tokens preceding the active-profile declaration. Schematically, changing P's grants changes the queried eligibility when P is active, but not when unchanged Q is active. We vary this declaration independently, testing both fact directions with the same candidate across routes. All consumers remain independent Handoff branches.

At the registered primary site, block 31, donor-correct judgments rise from 1 to 8/512 used cases (natural: 511/512). The gain and its used-minus-unused contrast are each +1.37 points, with adjusted 97.5\% intervals [0, 3.32]. Both lower bounds had to exceed zero, so the criterion fails. These contrasts score both routes against the used route's donor truth, which agrees with the unused route's gold in one fact direction and opposes it in the reverse. Table~\ref{tab:source} instead reports each route's own correctness (Appendix~\ref{app:source}).

\begin{table}[htbp]
\caption{\textbf{Source use determines the effect of an early patch.} Prespecified secondary Cross subset: 64 used and 64 unused cases in 16 reused permission worlds, matched across layers. Entries are correct counts out of 64. Select and Full use the active source; unused judgment gold is unchanged. Consumers are independent continuations.}
\label{tab:source}
\centering
\begin{tabular}{@{}lrrrr@{}}
\toprule
 & \multicolumn{2}{c}{Judgment} & \multicolumn{2}{c}{Action (used source)} \\
\cmidrule(lr){2-3}\cmidrule(l){4-5}
Intervention & Used & Unused & Select & Full \\
\midrule
Native & 0/64 & 64/64 & 8/64 & 5/64 \\
Patch, block 15 & 50/64 & 63/64 & 18/64 & 11/64 \\
Patch, block 31 & 0/64 & 64/64 & 9/64 & 6/64 \\
Patch, block 47 & 0/64 & 64/64 & 8/64 & 5/64 \\
Natural CF & 64/64 & 64/64 & 33/64 & 24/64 \\
\bottomrule
\end{tabular}

\end{table}

The predeclared block-15 check instead changes 50/64 used-source judgments toward the donor fact, with gains in every world and 25/32 in each direction (Table~\ref{tab:source}). One unused-source judgment is harmed. Matched block-31/47 target judgments do not change. Other judgments can also be harmed: block 15 reduces stable-eligible correctness from 122 to 120/128 and stable-ineligible correctness from 120 to 115/128.

Source editing also improves selection, but less consistently than judgment. Natural controls themselves reach only 33/64 correct selections and 24/64 complete actions despite perfect judgments. Thus the remaining action gap includes a capability limit, not only an intervention limitation. Earlier patches change more subsequent retained state, so this depth comparison does not uniquely locate a semantic representation.

\subsection{Target history leaves a learned-state bias in crossed selection}
\label{sec:bridge}
An exploratory composition uses the separate 16 discovery worlds. Let $I$ be the upper state (blocks 61--63) produced by the original writer on history A. AI/BI combine that same state with lower history A/B, without another residual write. Changing AI to BI raises Changed correctness to 49--55/64 across banks and menu families. Yet adding $I$ to B lower history (BA-to-BI) reduces Cross selection from 28 to 23/64 for Random-fact and from 21 to 16/64 for Path-cover, while old-ID responses rise from 2 to 11 and 5 to 15. Familiar selection remains 53/64 in both BI arms. Target history improves judgments without removing the learned state's crossed-selection bias (Table~\ref{tab:bridge}).

\section{Related work}
\label{sec:related}
\paragraph{Control, representation, and reuse.}
Function vectors transfer task behavior \citep{todd2024function}, but causal task control need not imply format-invariant representation \citep{opielka2026causality}. Closest to our crossing test, \citet{gao2026control} find extraction-index following after answer remapping; their separate behavioral evaluation also shows that MCQ and open-generation verdicts can differ. \citet{ma2026hidden} test reusable causal interfaces through forked operations and transplantation, while MQuAKE evaluates entailed consequences of knowledge edits \citep{zhong2023mquake}. Our retained-history interventions connect these concerns through factual recombination, independent condition queries, history composition, and source-use contrasts.

\paragraph{What an intervention establishes.}
DAS tests high-level and neural counterfactual agreement through aligned interchange interventions \citep{geiger2024alignments}. An intended output effect need not identify the proposed natural mechanism \citep{makelov2024illusion}; representational divergence can preserve observed behavior while changing pathways or behavior in other contexts \citep{grant2025divergence}. ReFT learns supervised hidden-state interventions in frozen models \citep{wu2024reft}. Unlike prompt-mimicking steering, which learns token-specific interventions from prompt-steered states \citep{heyman2026steer}, our fixed-site writer receives only dependent-ID supervision.

\paragraph{Bindings, probes, and downstream behavior.}
Binding-ID interventions can transfer across tasks \citep{feng2024binding}, and counterfactual embeddings test the behavioral role of syntactic representations \citep{tucker2021modified}. \citet{sun2026tool} decode tool-call dependencies from residual states, with reported patching that leaves calls unchanged. \citet{ravulapalli2026legible} use decoded bindings for error detection and residual repair in a controlled single-token task. We test reuse after successful free-generation control. Lookback mechanisms retrieve earlier bindings for belief questions \citep{prakash2025lookbacks}.

\section{Discussion and limitations}
\label{sec:discussion}
\paragraph{What successful control establishes.}
The learned writers improve familiar selection, but crossing removes the gain and verification benefits little. History composition improves judgments while retaining crossed-selection bias; the early source patch changes judgments according to source use. These comparisons show how interventions can differ in the computations their altered states support.

\paragraph{Which explanations remain?}
Target-ID supervision permits associations that fit training without preserving recomputation. Perfect-fit failures and old-ID attraction support this account, not a specific memorization circuit. The late writer leaves lower history untouched; early source patches change more retained state. Depth, representation and state coverage co-vary, leaving their causal contributions unresolved.

Verification changes computation and answer vocabulary. Truth-asymmetric winner accuracy and imperfect natural action selection further require consumer-specific capability controls: success on one query cannot stand in for competence on another.

\paragraph{Evaluation implications and scope.}
Evaluate edits with factual recombinations, independently elicited consequences, consumer-specific natural controls, and collateral checks. Task dependencies can guide source-use contrasts without assuming that the neural computation implements the task DAG.

We study one 27B model on procedural tasks, with privileged writer supervision and training-exposed Anchor menus. Interface/source tests reuse confirmation worlds; history composition uses separate discovery worlds. Source patches and writers differ in donor access, token support and depth, not just method. These experiments do not estimate deployed-agent error rates.

\section{Conclusion}
Successful residual writing can control a familiar target without supporting factual recomputation or independent verification. Composing learned state with target-condition history improves judgments without removing crossed-selection bias. Conversely, an early source patch changes judgments according to source use without ensuring complete actions. These findings distinguish selecting a target, using its governing condition, and completing the action as separate outcomes of an internal edit.

\label{maintextend}
\clearpage
\section*{Reproducibility statement}
The model revision, residual-writing objective and optimizer, generation settings, cohort definitions, and statistical procedure are specified in the paper and appendix. The reproducibility repository at \url{https://github.com/TiandaSun/residual-condition-reuse} contains frozen task definitions, scored outcomes, and scripts to reproduce statistics, tables and figures. For the source-intervention study, it also includes generated text for rescoring and bootstrap reproduction. These checks reuse saved outputs rather than rerun model inference; the package does not include all GPU tensors or a portable inference driver. Figure~\ref{fig:protocol} uses a shortened example; the frozen templates specify the full inputs.

\section*{Ethics statement}
The experiments use procedural explicit-rule episodes rather than private user interactions. Internal interventions can alter behavior outside the supervised target; full-action and stable-query outcomes are included to measure some of these effects. The results are an analysis of intervention behavior, not evidence that the procedure is safe for deployment or suitable for overriding real users' permissions.

\section*{LLM use statement}
Generative AI assisted research ideation, literature retrieval, experiment and analysis code development, procedural data construction, and manuscript drafting. References and numerical claims were checked against primary sources and recorded experiment artifacts. The authors remain responsible for the final scientific claims, artifact release, and submission. No AI-assisted proof of a mathematical theorem is claimed.

\bibliography{references}
\bibliographystyle{iclr2027_conference}
\clearpage
\appendix
\section{Runtime and optimization details}
\label{app:runtime}
The frozen model is \texttt{Qwen/Qwen3.5-27B}, revision
\begin{center}\small\texttt{fc05daec18b0a78c049392ed2e771dde82bdf654}.\end{center}
It has 64 blocks and hidden width 5,120. Runtime settings are BF16, SDPA attention, batch size one, greedy generation, no-thinking rendering, maximum 1,024 new tokens, and a 16,384-token context bound. The model prefills the entire supplied initial token sequence. The intervened \texttt{CX} token belongs to that sequence, rather than a newly sampled replacement for an initial decision. The block-output hook acts after the block's internal cache update. It is removed before future facts, future prefill, and generation.

For training view $v$, let $\ell_v(\delta)$ be mean cross-entropy over the 13 dependent-ID tokens: the shared \texttt{CX} token and twelve digit tokens. Model parameters are frozen. With $g_t=\frac18\sum_{v=1}^8\nabla_\delta\ell_v(\delta_t)$, optimization uses
\begin{equation}
\delta_{t+1}=\Pi_{\|\delta\|_2\leq R}\left(\delta_t-\eta_t\frac{g_t}{\|g_t\|_2}\right),
\qquad R=401.2637023925781.
\end{equation}
The zero-gradient case leaves the vector unchanged. The vector starts at zero in FP32; the realized activation is cast back to BF16. Step sizes are $R/12$ for updates 1--12, $R/48$ for 13--48, and $R/96$ for 49--96. Raw view gradients are averaged \emph{before} normalization, not applied as eight successive normalized updates. The endpoint is always step 96. Step 24 is a recorded intermediate checkpoint, not selected using evaluation results. No eligibility, independent-offer, reservation, or reader objective is included.

\section{Episode specification and illustrative example}
\label{app:example}
The model is given a condition-bearing initial history and a canonical initial action. Future menus describe dependent candidate IDs, cost and rule-relevant facts, together with independently constrained offers. The Select branch requests a \texttt{select\_followup} tool call with arguments \texttt{dependent\_id}, \texttt{independent\_id}, and \texttt{reservation\_id}. Query branches instead request an eligibility judgment in JSON. The public rules determine answers without consulting model outputs. The source menu selection and crossing satisfy frozen construction criteria; outcomes do not determine which cases enter confirmation.

The permission example in Figure~\ref{fig:protocol} comes from fresh world \texttt{B\_PERMISSION\_TRIANGLE-02}, direction Z0, Random-fact source view P1, Anchor. Candidate \texttt{CX829120987344} costs 33 and requires tags \texttt{TG61265761} and \texttt{TG78859500}. Profile A grants the former, whereas B grants both. Native A selects \texttt{CX277130775429}. The writer instead selects \texttt{CX829120987344}, with a complete action correct under B. An independent Changed branch about that same candidate returns \texttt{eligible:false}; Text update and Natural B return true. The figure aliases the tags as $p,q$ and the candidate as $s$. It omits the remaining candidates and independent offers for readability, and is not the literal full prompt. This post hoc example illustrates a verified same-entity mismatch; it does not make every aggregate query a judgment about the selected entity.

\paragraph{Crossed-fact example.}
The separate example in Section~\ref{sec:results} is fresh permission-path world 06, direction Z0, Random-fact source P0, Cross. Source training view V6 is crossed with facts from view V2. Alias $r$ is \texttt{CX875351644097}: its source cost is 24 and its required tags are both granted by B. After crossing, it costs 41 and requires \texttt{TG40446080} and \texttt{TG94769617}, both absent from B. Alias $s$ is \texttt{CX135949503030}, costing 78 with two B-granted tags; A's current winner is \texttt{CX103542360555}. The recorded writer call selects $r$, independent offer \texttt{IX953392826294}, and reservation \texttt{RX8829193039}. The latter two fields are correct, whereas the dependent target is wrong. This example has different entities from Figure~\ref{fig:protocol} and supplies no additional trial beyond the reported aggregate.

\section{Additional fresh-confirmation counts and statistics}
\label{app:stats}
The fresh generation seed is 202609240303. All 32 worlds, both directions, and both writer arms enter the primary analysis. ANCHOR and CROSS labels identify menu families, not independent cohorts. Of 64 optimizers per arm, 44 Random-fact and 39 Path-cover fit all eight dependent targets; only 17 per arm fit all complete tool actions. The complete-fit diagnostics are measured under free generation. Perfect target fit does not imply full-action fit because the other fields are not supervised.

\begin{table}[ht]
\caption{Additional fresh Cross outcomes. Each arm and family has 128 menus nested in 32 worlds. Old ID is the previously trained identifier, excluded from being the current B winner by the crossing construction.}
\centering\small
\begin{tabular}{llrrr}
\toprule
Bank & Arm & Select & Full & Old ID \\
\midrule
Random-fact & Native A & 37/128 & 22/128 & 6/128 \\
Random-fact & Writer & 30/128 & 15/128 & 33/128 \\
Random-fact & Text update & 65/128 & 37/128 & 7/128 \\
Random-fact & Natural B & 63/128 & 35/128 & 7/128 \\
Path-cover & Native A & 35/128 & 20/128 & 4/128 \\
Path-cover & Writer & 33/128 & 16/128 & 20/128 \\
Path-cover & Text update & 64/128 & 35/128 & 3/128 \\
Path-cover & Natural B & 58/128 & 32/128 & 6/128 \\
\bottomrule
\end{tabular}

\end{table}

\begin{table}[ht]
\caption{All five registered confirmation contrasts. Intervals use 100,000 joint PCG64 draws with seed 202609240399, linear percentile interpolation at 0.005 and 0.995, and eight worlds sampled with replacement within each of four fixed strata. $S$ and $D$ coincide because matched observed query gains are zero; this is not independent evidence counted twice.}
\centering\small
\begin{tabular}{lrr}
\toprule
Contrast & Estimate (pp) & 99\% interval (pp) \\
\midrule
S\_RANDOM & 65.63 & [57.03, 74.22] \\
S\_PATH & 63.28 & [53.13, 72.66] \\
D\_RANDOM & 65.63 & [57.03, 74.22] \\
D\_PATH & 63.28 & [53.13, 72.66] \\
K & 61.33 & [54.88, 68.75] \\
\bottomrule
\end{tabular}

\end{table}

The perfect-target-fit Cross subset contains 88 menus from 44 optimizers across 30 worlds for Random-fact, and 78 menus from 39 optimizers across 26 worlds for Path-cover. This post hoc descriptive subset does not affect inclusion in the primary analysis. The registered family remains the five contrasts on the entire fresh cohort. The discovery-world bridge uses exploratory 95\% marginal bootstrap intervals, not the fresh confirmation family; its bootstrap seed is 202609240400.

\begin{table}[ht]
\caption{Fresh cache-composition outcomes by rule family. All pools 512 menus; each family contains 256 menus from 16 worlds. The registered query contrast is BA minus AB, not BA minus AA, even though the latter gives the same pooled estimate here.}
\centering\small
\begin{tabular}{llrrrr}
\toprule
Family & State & Select & Full & Changed & Stable \\
\midrule
All & AA & 140/512 & 84/512 & 94/512 & 499/512 \\
All & AB & 139/512 & 81/512 & 94/512 & 499/512 \\
All & BA & 235/512 & 130/512 & 408/512 & 497/512 \\
All & BB & 234/512 & 131/512 & 407/512 & 498/512 \\
Permission & AA & 32/256 & 27/256 & 0/256 & 249/256 \\
Permission & AB & 32/256 & 27/256 & 0/256 & 249/256 \\
Permission & BA & 109/256 & 66/256 & 256/256 & 250/256 \\
Permission & BB & 109/256 & 66/256 & 256/256 & 250/256 \\
Time & AA & 108/256 & 57/256 & 94/256 & 250/256 \\
Time & AB & 107/256 & 54/256 & 94/256 & 250/256 \\
Time & BA & 126/256 & 64/256 & 152/256 & 247/256 \\
Time & BB & 125/256 & 65/256 & 151/256 & 248/256 \\
\bottomrule
\end{tabular}

\end{table}

The lower cache group contains 122/128 retained tensor components and approximately 95.32\% of bytes; the upper group contains six components. Components are individual cached tensors, not blocks. The upper group differs from the five components actually changed by the learned write. Continuation also preserves the required metadata. Direct Native A and Natural B continuations agree with cache AA and BB, respectively, in all $512\times3\times2=3{,}072$ comparisons (menus $\times$ consumers $\times$ native endpoints). Fresh BA/BB continuations agree on 504/512 Select, 511/512 Changed, and 511/512 Stable branches.

In the discovery-world composition, 2,046/2,048 query comparisons are token-identical. The 256 menu records already pool both source banks. Each is evaluated with both writer variants, including cross-bank evaluation, on two query types (Changed and Stable) and two state comparisons (AA with AI; BA with BI): $256\times2\times2\times2=2{,}048$. Table~\ref{tab:bridge} instead restricts each writer to its own bank, with 64 menus per family. The identity comparisons add learned upper state while holding lower history fixed, rather than comparing AI with BI.

\begin{table}[htbp]
\caption{Discovery-world history/writer composition. Each count is out of 64 own-bank menus from 16 discovery worlds. AI/BI contain the same learned upper state. Compare AI to BI for the history effect on Changed judgments, and BA to BI for adding learned state to B history. These are not the reused confirmation worlds in the source-patch study.}
\label{tab:bridge}
\centering\small
\begin{tabular}{llrrrrrr}
\toprule
Bank & Menu & \multicolumn{4}{c}{Select} & \multicolumn{2}{c}{Changed} \\ & & AA & BA & AI & BI & AI & BI \\
\midrule
Random-fact & Anchor & 15 & 28 & 59 & 53 & 12 & 50 \\
Random-fact & Cross & 14 & 28 & 12 & 23 & 11 & 51 \\
Path-cover & Anchor & 16 & 21 & 59 & 53 & 10 & 55 \\
Path-cover & Cross & 15 & 21 & 11 & 16 & 10 & 49 \\
\bottomrule
\end{tabular}

\end{table}

\section{Historical supervision and transfer diagnostics}
\label{app:discovery}
The three historical stages share 16 discovery worlds, four per pattern. The first compares four supervision arms; the second freezes the writers and tests transfer and readout; the third adds Anchor eligibility queries and natural-cache compositions. These are successive analyses of the same worlds, not three independent replications. Repeat duplicates one view, Label varies IDs/order, and Random-fact and Path-cover vary facts as described in the main text.

The first stage evaluates two held-out transformations. Both independently reassign cost and eligibility fields and shuffle display order. Joint also assigns new identifiers; Stable-ID-Joint preserves the original ID-to-row association, not the display position. The supervision comparison pools these two transformations, giving 384 held-out menus per arm.

\begin{table}[ht]
\caption{Historical step-96 supervision comparison. Held-out results pool Joint and Stable-ID-Joint menus. Training counts use each arm's eight logical views, including duplicates for Repeat. The Path-minus-Label held-out difference is 1.04 percentage points, with an exploratory 95\% interval [-1.04, 2.86]; it does not establish a graph-specific benefit.}
\centering\small
\begin{tabular}{lrrr}
\toprule
Arm & Training target & Held-out target & Held-out full \\
\midrule
Repeat & 256/256 & 77/384 & 40/384 \\
Label & 226/256 & 104/384 & 54/384 \\
Random-fact & 230/256 & 94/384 & 47/384 \\
Path-cover & 241/256 & 108/384 & 54/384 \\
Native A & -- & 91/384 & 53/384 \\
Text update & -- & 200/384 & 137/384 \\
\bottomrule
\end{tabular}

\end{table}

\begin{figure}[ht]
\centering
\begin{tikzpicture}[x=0.085cm,y=0.60cm,font=\normalsize]
\draw[->] (-88,0) -- (4,0);
\draw[gray!25] (-80,0) -- (-80,4.0);
\node[below] at (-80,0) {-80};
\draw[gray!25] (-60,0) -- (-60,4.0);
\node[below] at (-60,0) {-60};
\draw[gray!25] (-40,0) -- (-40,4.0);
\node[below] at (-40,0) {-40};
\draw[gray!25] (-20,0) -- (-20,4.0);
\node[below] at (-20,0) {-20};
\draw[gray!25] (0,0) -- (0,4.0);
\node[below] at (0,0) {0};
\draw[randomblue,line width=1.1pt] (-26.5625,3.55) -- (-7.8125,3.55);
\fill[randomblue] (-17.1875,3.55) circle (2.3pt);
\draw[randomblue,line width=1.1pt] (-78.1250,2.35) -- (-62.5000,2.35);
\fill[randomblue] (-70.3125,2.35) circle (2.3pt);
\draw[randomblue,line width=1.1pt] (-82.8125,1.15) -- (-65.6250,1.15);
\fill[randomblue] (-73.4375,1.15) circle (2.3pt);
\draw[pathorange,line width=1.1pt] (-23.4375,3.25) -- (-7.8125,3.25);
\fill[pathorange] (-15.6250,3.25) circle (2.3pt);
\draw[pathorange,line width=1.1pt] (-73.4375,2.0500000000000003) -- (-54.6875,2.0500000000000003);
\fill[pathorange] (-64.0625,2.0500000000000003) circle (2.3pt);
\draw[pathorange,line width=1.1pt] (-82.8125,0.85) -- (-68.7500,0.85);
\fill[pathorange] (-75.0000,0.85) circle (2.3pt);
\node[anchor=east] at (-90,3.4) {Order only};
\node[anchor=east] at (-90,2.2) {New IDs};
\node[anchor=east] at (-90,1.0) {Crossed facts};
\node[below] at (-42,-.6) {Change from familiar-menu target selection (percentage points)};
\fill[randomblue] (-71,4.55) circle (2.3pt);
\node[anchor=west] at (-68,4.55) {Random-fact};
\fill[pathorange] (-34,4.55) circle (2.3pt);
\node[anchor=west] at (-31,4.55) {Path-cover};
\end{tikzpicture}
\caption{Historical paired selection change from Anchor (59/64 for each bank). Intervals are exploratory 95\% pattern-stratified bootstraps over 16 worlds, not fresh confirmation intervals. Order-only, new-ID, and crossed-fact transformations separate changes that would be confounded by a single broad transfer score.}
\end{figure}
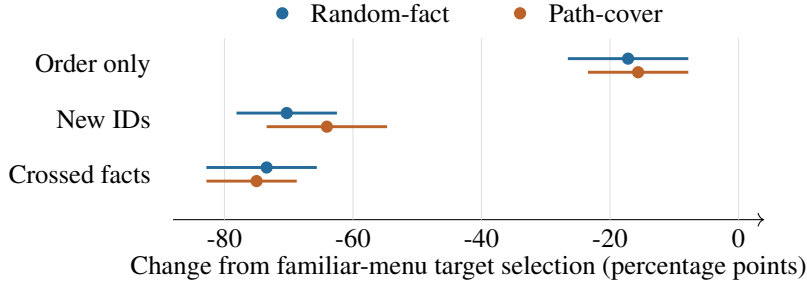

\begin{table}[ht]
\caption{Historical selection counts: 64 menus per row. Writer and Full report edited target-ID and full-action correctness, respectively; the other columns report target selection under each control.}
\centering\small
\begin{tabular}{llrrrrr}
\toprule
Writer bank & Menu & Writer & Full & Native A & Text & Natural B \\
\midrule
Random-fact & ANCHOR & 59/64 & 36/64 & 15/64 & 35/64 & 28/64 \\
Random-fact & ORDER-ONLY & 48/64 & 29/64 & 18/64 & 32/64 & 33/64 \\
Random-fact & ID-ONLY & 14/64 & 10/64 & 14/64 & 33/64 & 29/64 \\
Random-fact & CROSS & 12/64 & 6/64 & 14/64 & 31/64 & 28/64 \\
Path-cover & ANCHOR & 59/64 & 38/64 & 16/64 & 32/64 & 21/64 \\
Path-cover & ORDER-ONLY & 49/64 & 32/64 & 11/64 & 27/64 & 23/64 \\
Path-cover & ID-ONLY & 18/64 & 10/64 & 17/64 & 30/64 & 25/64 \\
Path-cover & CROSS & 11/64 & 8/64 & 15/64 & 26/64 & 21/64 \\
\bottomrule
\end{tabular}

\end{table}

\newpage
On historical Cross queries, Changed and Stable responses are each identical to Native A. The subsequent Anchor-query stage has Changed correctness 12/64 and 10/64, and Stable correctness 63/64 and 64/64, again with all matched query responses token-identical. Its cited 59/64 Anchor selections come from the earlier transfer stage rather than new Select executions. Historical cache Changed counts are 43, 44, 205, and 205 out of 256 for AA, AB, BA, and BB; these observations motivated, but are not pooled with, fresh confirmation.

Historical cache hashes show that an own-bank writer changes five of 128 retained components: block 61 recurrent state, block 62 convolution and recurrent states, and block 63 key/value tensors. Natural B changes 127/128. Mean edit-to-local-residual norm ratios are \RandomDose{} and \PathDose{}. Sparse support therefore should not be described as a small-amplitude perturbation.

\section{Eligibility-reader diagnostics}
\label{app:reader}
A frozen native-trained logistic reader estimates eligibility from the final-block candidate-object boundary in the Select prefill (C63), not the independent query branches. Its native dataset contains 256 worlds split into 192 training, 32 validation, and 32 qualification worlds, each with four menus, two conditions, and eight candidate positions. Features are standardized using training data. L2 regularization is selected by validation Brier score from $C\in\{0.01,0.1,1\}$; selected $C=0.1$. Validation temperature is approximately 1.134. Edited states do not select the reader.

Qualification balanced accuracy is \ReaderBA{}, versus \ReaderPermutationBA{} under label permutation. Aggregate performance masks weak sensitivity to condition-changing time pairs (Table~\ref{tab:readerqualification}). In fresh native-pair checks, both sides are correct for 97/384 time pairs and 837/1,024 permission pairs. The reader is consequently a diagnostic, not the basis for concluding that the condition is absent from all internal states.

\begin{table}[ht]
\caption{Historical native-reader qualification: both members of a condition-changing pair must be classified correctly. Candidate pairs share worlds and are not independent behavioral trials.}
\label{tab:readerqualification}
\centering\small
\begin{tabular}{lrr}
\toprule
Pattern & Both flip contexts correct & Rate \\
\midrule
Permission Path & 111/128 & 86.7\% \\
Permission Triangle & 106/128 & 82.8\% \\
Time One Middle & 8/32 & 25.0\% \\
Time Two Middle & 22/64 & 34.4\% \\
\bottomrule
\end{tabular}

\end{table}

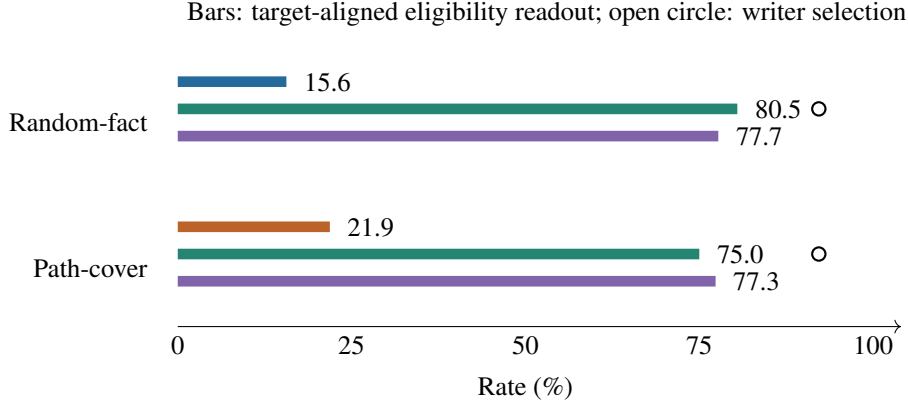
\begin{figure}[ht]
\centering
\begin{tikzpicture}[x=.092cm,y=1.20cm,font=\normalsize]
\node[anchor=east] at (-3,2.1999999999999997) {Random-fact};
\draw[randomblue,line width=4pt] (0,2.65) -- (15.6250,2.65);
\node[anchor=west,font=\normalsize] at (16.8250,2.65) {15.6};
\draw[textteal,line width=4pt] (0,2.3499999999999996) -- (80.4688,2.3499999999999996);
\node[anchor=west,font=\normalsize] at (81.6688,2.3499999999999996) {80.5};
\draw[naturalpurple,line width=4pt] (0,2.05) -- (77.7344,2.05);
\node[anchor=west,font=\normalsize] at (78.9344,2.05) {77.7};
\draw[black,thick,fill=white] (92.1875,2.3499999999999996) circle (2.5pt);
\node[anchor=east] at (-3,0.5999999999999998) {Path-cover};
\draw[pathorange,line width=4pt] (0,1.0499999999999998) -- (21.8750,1.0499999999999998);
\node[anchor=west,font=\normalsize] at (23.0750,1.0499999999999998) {21.9};
\draw[textteal,line width=4pt] (0,0.7499999999999998) -- (75.0000,0.7499999999999998);
\node[anchor=west,font=\normalsize] at (76.2000,0.7499999999999998) {75.0};
\draw[naturalpurple,line width=4pt] (0,0.44999999999999973) -- (77.3438,0.44999999999999973);
\node[anchor=west,font=\normalsize] at (78.5438,0.44999999999999973) {77.3};
\draw[black,thick,fill=white] (92.1875,0.7499999999999998) circle (2.5pt);
\draw[->] (0,-.05) -- (104,-.05);
\node[below] at (0,-.05) {0};
\node[below] at (25,-.05) {25};
\node[below] at (50,-.05) {50};
\node[below] at (75,-.05) {75};
\node[below] at (100,-.05) {100};
\node[below] at (50,-.5) {Rate (\%)};
\node[anchor=west,font=\normalsize] at (0,3.4) {Bars: target-aligned eligibility readout; open circle: writer selection};
\end{tikzpicture}
\caption{Historical Anchor readout, with 64 menus per bank. Bars show B-label classification accuracy at C63 on candidates whose eligibility changes between A and B; the threshold is 0.5. Blue/orange: writer; teal: Text update; purple: Natural B. Candidate means are averaged within menus, then within worlds, with equal weight on 16 worlds. Open circles show writer target-selection rates, a different outcome. Native qualification does not guarantee reader validity after a large residual change.}
\end{figure}

For reader probabilities $p_A,p_I,p_B$ under native, intervened, and natural-target histories, the diagnostic alignment gain is
\begin{equation}
G_{\rm JS}=\mathrm{JS}(\mathrm{Ber}(p_A),\mathrm{Ber}(p_B))
-\mathrm{JS}(\mathrm{Ber}(p_I),\mathrm{Ber}(p_B)).
\end{equation}
Changing-candidate rows are averaged within menus and then within worlds. This quantity compares Bernoulli predictions; natural-target reader predictions may themselves be wrong.

\newpage
\section{Frozen-writer interface diagnostics}
\label{app:interfaces}
The interface follow-up reuses the 32 confirmation worlds and all 128 step-96 endpoints without retraining. Its 63,488 free continuations comprise 6,144 reruns of the original interfaces, 8,192 new selection branches, and 49,152 Boolean branches. The reruns match the archived generations, including end-of-sequence behavior. These are repeated measurements on the same worlds. Every consumer starts independently from the same Handoff; no selected response is passed to another branch.

\paragraph{Tasks and matching.}
Selection requests the full action in plain JSON or a tool wrapper. Boolean branches return the candidate identifier and a truth value, as plain JSON or a \texttt{report\_predicate} tool call. Two fixed wordings are crossed with each wrapper. Within a wording, the wrapper comparison changes the formatting instruction; winner versus eligibility changes the question. The original-selection comparison additionally changes wording and is not a pure wrapper ablation.

Candidates include the A winner, B winner, and a sentinel: a deterministically chosen candidate eligible but not the winner under either condition. Its eligibility is always true and winner membership always false. The primary matched subset instead requires winner and eligibility truths to coincide under each condition and flip from A to B, excluding the sentinel. Candidate identity and labels are matched by public rules, without filtering on model outcomes. The questions can nevertheless differ in computational difficulty. Correctness requires both the correct candidate echo and Boolean, not merely a parseable response.

Every Anchor menu contributes one or two matched candidates. Random-fact has 100 menus with one and 28 with two; Path-cover has 96 with one and 32 with two. Thus each bank retains all 128 menus and all 32 worlds. Multiplying 156 and 160 candidates by two wordings gives the 312 and 320 paired questions per predicate and interface; no empty menu is excluded.

\paragraph{Estimands and intervals.}
Let $W$ and $E$ be writer-minus-Native gains on matched winner and eligibility questions. Candidate/wording averages are taken within menus, then within worlds, with equal weight on 32 worlds. The registered family contains eight effects: $W$ and $W-E$ for each of two banks and two Boolean interfaces. Joint pattern-stratified PCG64 bootstrap sampling uses 100,000 draws, seed 202609240499, and linear percentile bounds at 0.003125 and 0.996875. The resulting 99.375\% intervals implement approximate Bonferroni family control. None of the four combinations has positive lower bounds for both effects. This tests positive transfer; it is not a proof of equivalence to zero. Secondary selection, wrapper and attenuation intervals are descriptive 95\% intervals.

\begin{table}[ht]
\caption{Registered Boolean effects in percentage points. $W$ is winner gain; $E$ is eligibility gain. Each row reports two of the eight adjusted intervals.}
\centering\small
\begin{tabular}{llrr}
\toprule
Bank & Format & $W$ [99.375\%] & $W-E$ [99.375\%] \\
\midrule
Random-fact & JSON & -0.39 [-2.34, 1.17] & -0.39 [-2.34, 1.17] \\
Random-fact & Tool & 0.00 [0.00, 0.00] & -0.39 [-1.56, 0.00] \\
Path-cover & JSON & -1.17 [-3.13, 0.00] & -1.17 [-3.13, 0.00] \\
Path-cover & Tool & -0.39 [-1.56, 0.00] & -0.39 [-1.56, 0.00] \\
\bottomrule
\end{tabular}

\end{table}

\begin{table}[ht]
\caption{New selection interfaces: world-balanced effects and descriptive 95\% intervals (percentage points). Old-ID counts show Native-to-writer changes on Cross, out of 256 continuations per row.}
\centering\small
\begin{tabular}{llrrr}
\toprule
Bank & Format & Anchor gain [95\%] & Cross gain [95\%] & Old ID \\
\midrule
Random-fact & JSON & 60.55 [54.69, 66.41] & 1.95 [-4.30, 8.98] & 13$\to$51 \\
Random-fact & Tool & 62.50 [55.08, 69.53] & -1.56 [-7.81, 4.69] & 12$\to$50 \\
Path-cover & JSON & 57.42 [49.61, 65.23] & -4.69 [-8.59, -1.17] & 6$\to$33 \\
Path-cover & Tool & 57.42 [48.83, 66.02] & -2.34 [-7.03, 2.34] & 6$\to$30 \\
\bottomrule
\end{tabular}

\end{table}

\begin{table}[ht]
\caption{Directly paired original-selection-minus-winner attenuation, in percentage points, with descriptive 95\% intervals. The difference is estimated jointly, not inferred from separate significance decisions.}
\centering\small
\begin{tabular}{llr}
\toprule
Bank & Boolean format & Selection minus winner [95\%] \\
\midrule
Random-fact & JSON & 66.02 [59.38, 72.66] \\
Random-fact & Tool & 65.63 [59.38, 71.88] \\
Path-cover & JSON & 64.45 [57.03, 71.88] \\
Path-cover & Tool & 63.67 [56.25, 71.09] \\
\bottomrule
\end{tabular}

\end{table}

Random-fact selects the correct dependent ID on 220/256 JSON and 224/256 tool-wrapped continuations; Path-cover reaches 210/256 and 214/256. Full-action correctness is 129/256 and 127/256 for Random-fact, and 122/256 and 119/256 for Path-cover, in the same format order.

Across the 12,288 paired Boolean branches, 12,257 complete generations are identical between Native and writer; the 31 changes include six correctness gains, 21 losses, and four pairs wrong under both. Sixteen writer responses echo a wrong candidate ID despite parseable JSON; six selection responses contain schema-valid strings with invalid candidate-ID format. These outcomes are retained, not removed as parser failures.

\paragraph{Capability and illustrative instances.}
The main-text truth split exposes weak natural target-true winner verification. A post hoc screen identifies 27 permission records (14 Random-fact, 13 Path-cover) in 12 worlds. A record fixes a world, condition direction, writer bank and Anchor menu. For the same target candidate, all five writer selection variants choose B's winner; Native A answers all eight Boolean questions correctly under A, and Natural B and Text each answer all eight correctly under B. All eight writer judgments are wrong under B. Fourteen records also have all five complete actions correct. Success of the controls and failure of the writer judgments are both selection criteria: these examples do not estimate a failure rate among all capable cases.

One verified instance is permission-path world 01, direction Z0, Random-fact source P0, Anchor. The B winner \texttt{CX432563164842} costs 28 and requires two B-only permissions, \texttt{TG30135501} and \texttt{TG00615576}; the A winner \texttt{CX793777112318} costs 62 and requires shared permissions. In all five variants, the writer selects the B winner, independent offer \texttt{IX693040549940}, and reservation \texttt{RX8072357918}. All eight independent writer judgments return false, as does Native A, whereas Natural B and Text update correctly return true. Anchor remains training-exposed. This is not a continuous conversation in which the model first chooses and then denies its choice.

\paragraph{Passive logit diagnostics.}
We record existing forward-pass logits at the Boolean-answer boundary, without rerunning the model with a forced prefix. Of 49,152 opportunities, 49,136 reach the boundary; 12,272 writer/Native pairs have comparable prefixes, and 8,697 have nonzero margin changes. Sixteen noncomparable pairs involve the writer's wrong candidate echo and remain in the behavioral analysis. The true/false tokens carry at least 0.99920 of full-vocabulary mass. We normalize over those two tokens to obtain $P(\mathrm{true})$. For target truth $b$, the target-oriented change is $(2b-1)\Delta P(\mathrm{true})$.

For matched Anchor questions with comparable prefixes, query-weighted mean target-oriented changes range from -0.226 to -0.137 percentage points for winner verification and from 0.102 to 0.122 for eligibility. Mean absolute changes range from 0.762 to 1.158 points. Output probabilities can therefore change even when the generated answer is unchanged. These small shifts do not establish condition reuse or identify its mechanism. They are not the world-balanced primary estimand.

\section{Source-fact interventions and routing controls}
\label{app:source}
\paragraph{Construction and cohort.}
We reuse all 16 permission worlds from fresh confirmation and all 128 Random-fact menu records, without filtering on model behavior. Each world contains two profiles, indexed by their condition references rather than display order. For source $s$, fact world F0 is the original world; F1 replaces only that profile's granted-tag array with the other profile's original array. Profile identity, position, other fields, menus and rules remain fixed. Both F0-to-F1 and F1-to-F0 directions are evaluated. Independently, route $r$ declares which profile is active. The edited source is used if $r=s$ and unused otherwise. A public solver verifies the canonical initial action in every combination.

The Changed candidate is selected by a frozen hash rule from candidates whose eligibility differs between the two original profiles, and held fixed across sources, directions and routes. Separate stable-eligible and stable-ineligible candidates have unchanged truth. Select still requests all three tool arguments. Every consumer independently resumes the parent Handoff, so joint judgment/selection counts describe corresponding branches rather than sequential decisions.

\paragraph{Source support and native controls.}
The patch replaces the entire token span covering the selected \texttt{granted\_tags} array with natural donor residual rows at a fixed block output. The 51- or 61-token spans cover array delimiters and fixed boundary punctuation, not profile IDs or the later routing declaration. Donor/recipient token sequences are aligned and identical outside the span. Source activations are captured before routing and are identical across the two routes; neither the future menu nor the consumer question is available at capture time. Natural counterfactual controls instead prefill the entire donor fact world at the same route.

Identity at block 31 replaces each span with its own native activations. All 4,096 matched Native/Identity continuations have identical generated tokens. Genuine donor patches have nonzero differences, exactly replace the nominated rows, and leave non-span rows unchanged at the hook. The hook acts after its block updates its own cache and is removed before future computation. Consequently, retained state at or below that block stays recipient-identical, while later layers process the altered source (Table~\ref{tab:source_support}). This is a source-span intervention, not a single-token learned vector or a localized eligibility neuron.

\begin{table}[htbp]
\caption{Persistent-state changes under source patches. Components are convolution, recurrent, key or value tensors. Earlier patches alter more subsequent state; depth and coverage are not independently manipulated.}
\label{tab:source_support}
\centering\small
\begin{tabular}{@{}lrrrrr@{}}
\toprule
Intervention & Layers & Components & Conv & Recurrent & K/V \\
\midrule
Patch, block 15 & 16--63 & 95 & 35 & 36 & 12/12 \\
Patch, block 31 & 32--63 & 63 & 23 & 24 & 8/8 \\
Patch, block 47 & 48--63 & 31 & 11 & 12 & 4/4 \\
Writer, block 60 & 61--63 & 5 & 1 & 2 & 1/1 \\
Patch31 + writer & 32--63 & 63 & 23 & 24 & 8/8 \\
\bottomrule
\end{tabular}

\end{table}

\paragraph{Registered target-event effects.}
For each paired source/menu/direction, let $t$ be donor eligibility when the source is used, and let $q_t(y)$ indicate a schema-valid answer with the correct candidate echo and truth value $t$. The same $t$ is used under both routes. For the block-31 patch $P$ and identity control $I$, define
\begin{equation}
 A=\mathbb{E}[q_t(P_{\mathrm{used}})-q_t(I_{\mathrm{used}})],\quad
 U=\mathbb{E}[q_t(P_{\mathrm{unused}})-q_t(I_{\mathrm{unused}})],\quad J=A-U.
\end{equation}
Unlike the accuracy columns in Table~\ref{tab:source}, $U$ measures the same target event as $A$, not correctness under the unused route's own gold. Within worlds, sources/directions and the eight original menu records receive equal weight, followed by equal world weighting. There are 512 used/unused pairs nested in 16 worlds. The two co-primary effects are $A$ and $J$; both adjusted lower bounds must exceed zero. We use 100,000 PCG64 bootstrap draws, seed 202609250199, sampling eight worlds with replacement in each permission-pattern stratum. Percentile 97.5\% intervals apply approximate Bonferroni correction; descriptive intervals are 95\%. Natural effects replace patch/identity with natural counterfactual/Native using the same target event.

\begin{table}[htbp]
\caption{Registered block-31 source effects and natural controls in percentage points. Neither co-primary adjusted lower bound is positive. The observed unused-source zero is not a population equivalence result.}
\centering\small
\begin{tabular}{@{}lrrr@{}}
\toprule
Contrast & Effect (pp) & 95\% interval & 97.5\% interval \\
\midrule
Patch: used ($A$) & +1.37 & [0.00, 3.13] & [0.00, 3.32] \\
Patch: unused ($U$) & +0.00 & [0.00, 0.00] & --- \\
Patch: selectivity ($J$) & +1.37 & [0.00, 3.13] & [0.00, 3.32] \\
Natural: used & +99.61 & [98.83, 100.00] & --- \\
Natural: unused & +0.00 & [0.00, 0.00] & --- \\
Natural: selectivity & +99.61 & [98.83, 100.00] & --- \\
\bottomrule
\end{tabular}

\end{table}

The seven block-31 gains are valid opposite-to-target Boolean changes, not invalid-output repairs, and occur in only three worlds. Natural donor judgment correctness is 511/512; Native is 1/512. All unused-source Changed judgments remain correct (512/512), but stable-ineligible correctness falls from 990 to 982/1,024 with seven gains and fifteen losses. Thus unchanged target queries do not imply unchanged collateral behavior.

\newpage
\paragraph{Prespecified layer comparison.}
Blocks 15 and 47 are fixed secondary checks, not selected by an adaptive layer search. Their subset fixes Cross, original direction Z0 and source menu P0, retaining every permission world, both sources, routes and fact directions. Each layer has 64 used and 64 unused cases; block 31 in Table~\ref{tab:source} is restricted to these exact cases, not its entire primary cohort. All 50 block-15 judgment gains come from valid Boolean changes. Every world contributes one to four gains out of four used-source queries. Table~\ref{tab:source_strata} separates direction, target truth, source identity and physical display position. These overlapping strata are descriptive, not independent replications.

\begin{table}[htbp]
\caption{Block-15 used-source judgment gains on the fixed secondary subset. Rows partition the same 64 cases in different ways; source identity is distinct from display position.}
\label{tab:source_strata}
\centering\small
\begin{tabular}{@{}llrrr@{}}
\toprule
Stratum & Group & Correct & Gains & Losses \\
\midrule
Direction & $F_0\to F_1$ & 25/32 & 25 & 0 \\
Direction & $F_1\to F_0$ & 25/32 & 25 & 0 \\
Target truth & False & 28/32 & 28 & 0 \\
Target truth & True & 22/32 & 22 & 0 \\
Source identity & S0 & 22/32 & 22 & 0 \\
Source identity & S1 & 28/32 & 28 & 0 \\
Source display & First & 19/32 & 19 & 0 \\
Source display & Second & 31/32 & 31 & 0 \\
Pattern & Path & 26/32 & 26 & 0 \\
Pattern & Triangle & 24/32 & 24 & 0 \\
\bottomrule
\end{tabular}

\end{table}

The single unused-source harm is a false-to-true response in the forward fact direction. Across both routes, stable-eligible correctness falls 122 to 120/128 (two losses); stable-ineligible falls 120 to 115/128 (seven gains, twelve losses). Used-source target selection has eleven gains and one loss, and donor-legal selection rises 41 to 57/64. Of the 50 correct edited judgments, 16 are paired with a correct target selection in the independent branch and 34 are not. This is not a count of selecting the same candidate judged ineligible: that exact inconsistency occurs once among the 64 used cases.

\paragraph{An illustrative same-candidate case.}
In permission-triangle world 02, original Z0, Random-fact P0, Cross, source 1 is active and the reverse fact change removes required grants from candidate \texttt{CX203840942635}. Block 15 changes its judgment from true to false, but the separate Select branch retains that candidate. Natural counterfactual history answers false and chooses the donor-optimal \texttt{CX961577507878}. When source 1 is unused, all tested arms answer true and select \texttt{CX203840942635}, as the unchanged gold requires. This post hoc case illustrates a cross-branch mismatch, not its population frequency.

\paragraph{Composition with the frozen writer.}
The additional composition uses only block 31, not the successful block-15 support. It retains Cross records whose active route matches the original writer bank and loads the 32 original Random-fact step-96 vectors. The source patch occurs first; the same historical vector is then added at the initial block-60 identifier site. This preserves the vector, not its resulting upper state, because the source patch changes the residual entering it.

\begin{table}[htbp]
\caption{Source-patch/writer composition on 128 used and 128 unused cases. Judgments and selections are independent. Old ID is the original training target; donor correctness concerns the current source-fact counterfactual. The source patch here is block 31, not block 15.}
\centering\small
\begin{tabular}{@{}llrrrr@{}}
\toprule
Intervention & Source & Judgment & Select & Full & Old ID \\
\midrule
Native & Used & 0/128 & 17/128 & 14/128 & 10/128 \\
Patch31 & Used & 2/128 & 18/128 & 14/128 & 12/128 \\
Writer & Used & 0/128 & 11/128 & 7/128 & 31/128 \\
Patch31 + writer & Used & 2/128 & 12/128 & 8/128 & 29/128 \\
Native & Unused & 128/128 & 69/128 & 48/128 & 9/128 \\
Patch31 & Unused & 128/128 & 69/128 & 48/128 & 8/128 \\
Writer & Unused & 128/128 & 35/128 & 29/128 & 34/128 \\
Patch31 + writer & Unused & 128/128 & 33/128 & 27/128 & 35/128 \\
\bottomrule
\end{tabular}

\end{table}

\newpage
The writer retains old-ID attraction, while the two block-31 donor-correct judgments persist in composition. This does not test whether a successful block-15 correction survives the writer. In the 128 unused cases, judgments are correct throughout because their gold is unchanged. Across all 256 composition cases, donor and original opposite-condition winners coincide in 64.

Excluding these 64 cases, Native/Patch/Writer/Combined old-ID counts are 14/15/47/47 out of 192 and donor-correct selections are 75/75/41/39.

\paragraph{Execution checks.}
The source study completes 19,456 distinct experimental generations: 16,384 core, 1,024 boundary and 2,048 composition outputs. All reach EOS and satisfy the frozen schema; four composition selections nevertheless contain an invalid candidate-ID string, retained as strict failures. Archived checks cover matched source tokens, complete hybrid state, 48 original-state save/reload checks and all 32 historical endpoints. Additional zero-write and hook-order checks were completed during the experiment, not all before it; they leave the reported set of experimental outputs unchanged. The 112 engineering generations are separate. Raw-output rescoring, pairing and primary-bootstrap reproduction are recorded independently of model execution.

\section{Reproducibility checks and measurement counts}
We archive generation outputs, scored rows, frozen protocols, model settings and source hashes separately for each experimental stage. The transfer, query/cache, fresh-confirmation, interface and source-intervention audits rescore 5,952, 4,864, 19,456, 63,488 and 19,456 decoded outputs using pinned scoring implementations. The earlier supervision stage is authenticated and reaggregated without a new raw-output rescore. Confirmation, interface and source-intervention research audits also reproduce their registered bootstraps. These checks reuse the experimental outputs; they are not new model runs or validation by an independent parser.

The 19,456 confirmation-stage continuations comprise 7,680 fresh writer/consumer branches, 6,144 fresh cache branches, 1,024 training-fit evaluations and 4,608 discovery-world composition branches. Interface and source-stage breakdowns appear in Appendices~\ref{app:interfaces} and~\ref{app:source}. Generations are repeated measurements, not independent examples. The confirmation population remains 32 fresh worlds, with 16 discovery worlds reported separately; source interventions reuse the 16 confirmation permission worlds. Numeric tables and plots are generated from audited data rather than pooling these stages as new populations.

\end{document}